\documentclass{article}
\usepackage{spconf,amsmath,graphicx}
\usepackage{multirow}
\usepackage{array}
\usepackage{cite}
\usepackage{amssymb}
\usepackage{mathrsfs}
\usepackage{blindtext}
\usepackage{algpseudocode}
\usepackage{algorithm}
\usepackage{multirow}
\usepackage{bm}
\usepackage{booktabs}
\usepackage{url}
\usepackage[super]{nth}
\title{EURO: ESPnet Unsupervised ASR Open-source Toolkit}
\name{
\begin{tabular}{c}Dongji Gao$^1$\sthanks{Equal Contribution.}, Jiatong Shi$^{2*}$, Shun-Po Chuang$^{3}$, Leibny Paola Garcia$^1$, Hung-yi Lee$^{3}$, \\
Shinji Watanabe$^2$, Sanjeev Khudanpur$^1$
\end{tabular}
}
\address{$^1$Johns Hopkins University, $^2$Carnegie Mellon University, $^3$ National Taiwan University \\
\small{\texttt{\{dgao5,lgarci27,khudanpur\}@jhu.edu}, 
\texttt{\{jiatongs,swatanab\}@cs.cmu.edu},} \\
\small{\texttt{\{f04942141,hungyilee\}@ntu.edu.tw}}
}

\begin{document}
\ninept
\maketitle
\begin{abstract}
This paper describes the ESPnet Unsupervised ASR Open-source Toolkit (EURO), an end-to-end open-source toolkit for unsupervised automatic speech recognition (UASR).
EURO adopts the state-of-the-art UASR learning method introduced by the Wav2vec-U, originally implemented at FAIRSEQ, which leverages self-supervised speech representations and adversarial training. In addition to wav2vec2, EURO extends the functionality and promotes reproducibility for UASR tasks by integrating S3PRL and k2, resulting in flexible frontends from 27 self-supervised models and various graph-based decoding strategies. EURO is implemented in ESPnet and follows its unified pipeline to provide UASR recipes with a complete setup. This improves the pipeline's efficiency and allows EURO to be easily applied to existing datasets in ESPnet.
Extensive experiments on three mainstream self-supervised models demonstrate the toolkit's effectiveness and achieve state-of-the-art UASR performance on TIMIT and LibriSpeech datasets. 
EURO will be publicly available at \texttt{https://github.com/espnet/espnet}, aiming to promote this exciting and emerging research area based on UASR through open-source activity.

%The abstract should appear at the top of the left-hand column of text, about
%0.5 inch (12 mm) below the title area and no more than 3.125 inches (80 mm) in
%length.  Leave a 0.5 inch (12 mm) space between the end of the abstract and the
%beginning of the main text.  The abstract should contain about 100 to 150
%words, and should be identical to the abstract text submitted electronically
%along with the paper cover sheet.  All manuscripts must be in English, printed
%in black ink.
\end{abstract}
\section{Introduction}
\label{sec:intro}
Over the past decade, end-to-end (E2E) supervised ASR has achieved outstanding improvements. These achievements keep pushing the limit of ASR performance in terms of word error rate (WER). 
However, training the state-of-the-art model still heavily relies on a reasonable amount of annotated speech \cite{contextnet, gulati2020conformer, guo2021recent}. Unfortunately, labeled data is quite limited for most of the 7000 languages worldwide \cite{grenoble2011handbook}. The fast development of self-supervised learning (SSL) could mitigate the issue to some extent by leveraging unlabeled data. This paradigm first learns the speech representations from raw audio and then fine-tunes the model on limited transcribed speech data~\cite{wav2vec2, hubert,wavlm}, resulting in reducing the need for annotated speech. 

However, as there is still a need for transcribed data for downstream model training, it is difficult to directly apply the system to all languages, especially those endangered languages that are extremely difficult to obtain data \cite{michaud2014towards, shi2020leveraging}. Unsupervised ASR could be one possible direction to solve the problem where the model can be trained with more accessible unpaired speech and text data. 
% The idea of unsupervised learning is quite intuitive. As we know, humans often acquire knowledge with little or no supervision.  The structure of information is mostly learned from trial and error.

Wav2vec-U is the state-of-the-art UASR framework. It utilizes both SSL (i.e., wav2vec2.0) and adversarial training for the UASR task\cite{wav2vecu}. The framework is shown working not only in English, where the wav2vec2.0 \cite{wav2vec2} is trained on but also in several other mainstream languages~\cite{librispeech, timit, mls}, as well as low-resource languages~\cite{swahili}. This finding reveals the possibility of utilizing unsupervised learning for more languages. %Notably, the work is open-sourced in Faiseq \cite{fairseq}.

The authors of wav2vec-U have released their code in FAIRSEQ \cite{fairseq}, which greatly improves reproducibility. The implementation mainly consists of 3 steps: data preparation, generative adversarial training (GAN)~\cite{gan}, and iterative self-training~\cite{iter} followed by Kaldi LM-decoding~\cite{kaldi}. 
Along with this reproducibility direction, we develop an unsupervised ASR toolkit named ESPnet Unsupervised ASR Open-source toolkit (EURO).
EURO complements the original FAIRSEQ implementation with more efficient multi-processing data preparation, flexible choices over different SSLs, and large numbers of ASR tasks through ESPnet~\cite{espnet}. EURO also integrates a weighted finite-state transducers (WFST) decoder using the k2~\cite{k2} toolkit for word-level recognition.
k2 is the updated version of the popular ASR toolkit Kaldi~\cite{kaldi}. It seamlessly integrates WFST and neural models implemented in PyTorch~\cite{pytorch} by supporting automatic differentiation for finite state automaton (FSA) and finite state transducer (FST), which are commonly used in ASR as a natural representation of the model's architecture ~\cite{wfst}. In EURO, k2 provides a compact WFST structured and efficient algorithm for decoding. With these advantages, EURO can considerably benefit the UASR study for the speech community, together with the FAIRSEQ UASR implementation.
%However, because of some design issues, it cannot offer enough flexibility for usage, including inefficient data preparation, pre-defined SSL feature supports, and difficulties in transferring to new datasets.

%In this paper, we introduce an unsupervised ASR toolkit named ESPnet Unsupervised ASR Open-source toolkit (EURO).
%EURO follows the Wav2vec-U framework~\cite{wav2vecu}
%but extends the functionality by integrating S3PRL, a toolkit for SSL, to allow flexible choices over different SSL models. The consistent pipeline with ESPnet makes EURO easily to be applied to datasets in ESPnet.
This paper first introduces the toolkit and its features. Then, we conduct experiments that explore mainstream self-supervised models as speech feature extractors for UASR. Finally, we provide details of the hyperparameters of our experiments.
% In addition, this paper introduces two auxiliary objectives to increase the stability of the training.
%These guidelines include complete descriptions of the fonts, spacing, and
%related information for producing your proceedings manuscripts. Please follow
%them and if you have any questions, direct them to Conference Management
%Services, Inc.: Phone +1-979-846-6800 or email
%to \\\texttt{papers@2021.ieeeicassp.org}.

\section{Related works}
\label{sec:related works}
%Current research on the UASR task mainly focuses on phoneme-level sequence prediction \cite{aldarmaki2022unsupervised} using unparallel audio and phonemicized text. It learns to map discrete acoustic representations into phonemes by fitting the predicted phoneme sequences into the structure of phonemicized text sequences (e.g., phoneme language model). 
This section briefly compares the framework of EURO to wav2vec-U.
As summarized in Table~\ref{tab: comparison}, EURO provides more flexible choices of SSL model as the acoustic feature extractor by integrating with the S3PRL toolkit~\cite{s3prl}. With the comprehensive pipeline in the template, EURO enjoys a fast adoption to various datasets with a limited data preparation effort (less than 20 lines of code for a minimum runnable solution). 
%EURO also includes a self-implemented decoder which eliminates the need for external dependency, and 
Meanwhile, all the data preparation stages in EURO are designed to enable computing in parallel, which greatly minimized the preprocessing time compared to wav2vec-U.
Besides of WFST decoder introduced in Sec.~\ref{sec:intro}, EURO provides a self-implemented decoder to eliminate external dependencies. 
\begin{table}[t]
    \caption{Framework comparison of EURO with wav2vec-U}
    \vspace{1mm}
    \centering
    \resizebox{\columnwidth}{!}{
    \begin{tabular}{ll|cc}
    \toprule
       Features & Details  & Fairseq & EURO \\
    \midrule
        \multirow{1}{*}{Model} & Frontend & 1 SSL  & 27 SSLs in \cite{s3prl} \\
%         & \multirow{2}{*}{Arch.} & wav2vec-U  \cite{wav2vecu} & \multirow{2}{*}{wav2vec-U \cite{wav2vecu}} \\
%        & & wav2vec-U2 \cite{liu2022towards} & \\
         \midrule          
        \multirow{3}{*}{Efficiency} & Prepare & single-thread & multi-process \\
        & Train & multi-GPU & multi-GPU \\
        & Decode & single-thread & multi-process \\
        \midrule
        \multirow{2}{*}{Decoding} & Prefix & by FlashLight \cite{kahn2022flashlight} & self-implemented \\
        & WFST & by PyKaldi \cite{can2018pykaldi} & by k2\\

    \bottomrule
    \end{tabular}}
    \label{tab: comparison}
    \vspace{-10pt}
\end{table}

\section{Functionalities of EURO}
\label{sec:euro}

\begin{figure}[]
\begin{minipage}[b]{0.56\linewidth}
    \centering
    \centerline{\includegraphics[width=5.5cm]{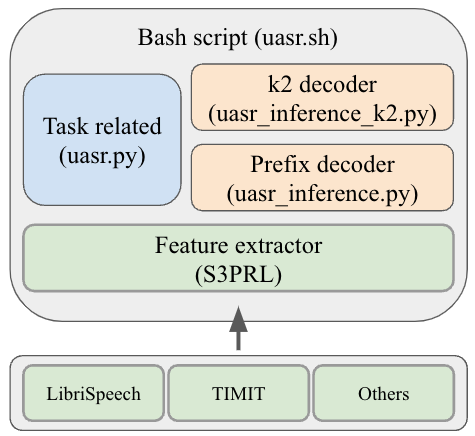}}
    \caption{Architecture of EURO}
    \label{fig:architecture}
\end{minipage}
\hfill
\begin{minipage}[b]{0.4\linewidth}
    \centering
    \centerline{\includegraphics[width=3.2cm]{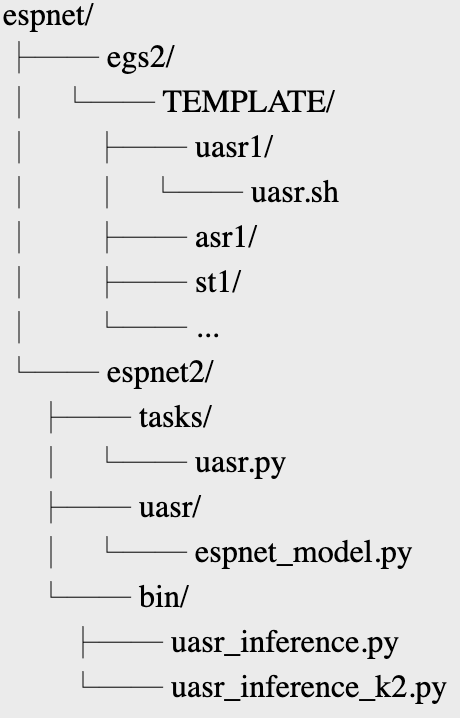}}
    \caption{Directory of EURO}
    \label{fig:tree}
\end{minipage}
\vspace{-15pt}
\end{figure}

Fig.~\ref{fig:architecture} shows the architecture of EURO. Similar to other ESPnet tasks, EURO includes two major components: a Python library of network training/inference and a collection of recipes for running complete experiments for a number of datasets. The library is built upon PyTorch, while the recipes offer all-in-one style scripts that follow the data format style in Kaldi \cite{kaldi} and ESPnet \cite{espnet}. In addition, a WFST decoder is included to perform word-level recognition.

\subsection{Models}
\label{ssec: models}

As discussed in Sec.~\ref{sec:related works}, we follow previous works in using adversarial training to achieve UASR in EURO. To be specific, we extend the Wav2vec-U framework into our implementation. 

Given a spoken utterance $\mathbf{X} \in \mathcal{D}_{\text{speech}}$, we first extract the speech representation by using a speech SSL model as the feature extractor $f(\cdot)$, resulting in a sequence of hidden representations $\mathbf{H}$. Then, the sequence $\mathbf{H}$ is passed into a preprocessor $m(\cdot)$ to form a segmented feature sequence $\mathbf{S}$, which is used for the generative adversarial network (GAN). The preprocessor $m(\cdot)$ includes three steps, including adjacent clustering pooling, principle component analysis (PCA) dimension reduction, and mean pooling. The adjacent clustering pooling utilizes the K-Means cluster IDs from the input feature $\mathbf{H}$ as guidance to merge the adjacent feature frames.

The network is mainly trained with a GAN-based loss and some auxiliary supporting losses. Given the segmented feature $\mathbf{S}$ and an unpaired phonemicized text sequence $\mathbf{Y}_{u} \in \mathcal{D}_{\text{text}}$, the framework includes a generator $\mathcal{G}$ and a discriminator $\mathcal{C}$ as the classic GAN framework. The generator $\mathcal{G}$, which serves as the ASR model, transcribes $\textbf{S}$ into a phoneme sequence $\mathbf{P}$ and the discriminator $\mathcal{C}$ tries to distinguish $\mathbf{Y}_u$ from the generated phoneme sequence $\mathbf{P}$. The GAN-based loss $\mathcal{L}_{\text{GAN}}$ is as follows:
\begin{equation}
    \mathcal{L}_{\text{GAN}} = \min_{\mathcal{G}} \max_{\mathcal{C}} \mathbb{E}_{\mathbf{Y}_u} [\log \mathcal{C}(\mathbf{Y}_u)] - \mathbb{E}_{\mathbf{H}} [\log(1 - \mathcal{C(\mathcal{G}(\mathbf{S}))})].
\end{equation}
\noindent
To stabilize the training, three auxiliary losses are also proposed, including (a) a gradient penalty loss $\mathcal{L}_{\text{gp}}$ to sample the mixing rate of real and fake input for different steps:

\begin{equation}
    \mathcal{L}_{\text{gp}} = \underset{\mathbf{S}, \mathbf{Y}_u, \alpha \sim U(0, 1)}{\mathbb{E}} [(||\nabla \mathcal{C}(\alpha \mathcal{G}(\mathbf{S}) + (1 - \alpha) \mathbf{Y}_u)|| - 1)^2],
\end{equation}
\noindent where $\alpha$ is the mixing weight sampled from a uniform distribution~\cite{gulrajani2017improved}. (b) a smoothness penalty to penalize inconsistent phoneme prediction between adjacent segments:
\begin{equation}
    \mathcal{L}_{\text{sp}} = \underset{(p_n, p_{n+1}) \in \mathcal{G}(\mathbf{S})}{\sum} ||p_n - p_{n+1}||^2,
\end{equation}
\noindent where $p_n \in \mathbf{P}$ is the generator output distribution at the $n$'s segment. (c) a phoneme diversity loss to prevent the generator $\mathcal{G}$ from generating the same phoneme all the time:
\begin{equation}
    \mathcal{L}_{\text{pd}} = -\sum_{n} \mathrm{Entropy}_{\mathcal{G}}(\mathcal{G}(\mathbf{S})),
\end{equation}
\noindent that is defined by the entropy of the average generator output distribution over every frame. The final loss is defined as
\begin{equation}
    \mathcal{L} = \mathcal{L}_{\text{GAN}} + \lambda \mathcal{L}_{\text{gp}} + \gamma \mathcal{L}_{\text{sp}} + \eta \mathcal{L}_{\text{pd}},
\end{equation}
\noindent where $\lambda, \gamma, \eta$ are the weights for each term.

\subsection{Frontend}
\label{ssec:frontend}

One of the major benefits of EURO compared to wav2vec-U is its tight integration with S3PRL\footnote{\url{https://github.com/s3prl/s3prl}}, a toolkit for speech/audio self-supervised models, to avoid manually managing and switching between different SSL models. 
S3PRL supports various SSL models as a general toolkit. 
S3PRL has kept up-to-date with the latest SSL models in the speech and audio domain. Based on the integration with S3PRL, by simply changing \textbf{one line} of the configuration, EURO can utilize up to \textbf{27} speech and audio SSLs with more than \textbf{70} of their variants.\footnote{The number is recorded in Oct. 2022.}

\subsection{Decoding}
\label{ssec: decoding}

EURO offers two methods for decoding, including a self-implemented prefix beam search method and a graph-based search method using k2.

The prefix beam search utilizes the same decoding process as the CTC prefix decoding \cite{alex2008supervised, watanabe2017hybrid}, but without the blank symbols. Similar to other ESPnet tasks, the decoding can be integrated with phoneme-level language models (LM) from both n-gram LMs and neural LMs.

\begin{figure}[]
\begin{minipage}[b]{0.48\linewidth}
    \centering
    \centerline{\includegraphics[width=4.0cm]{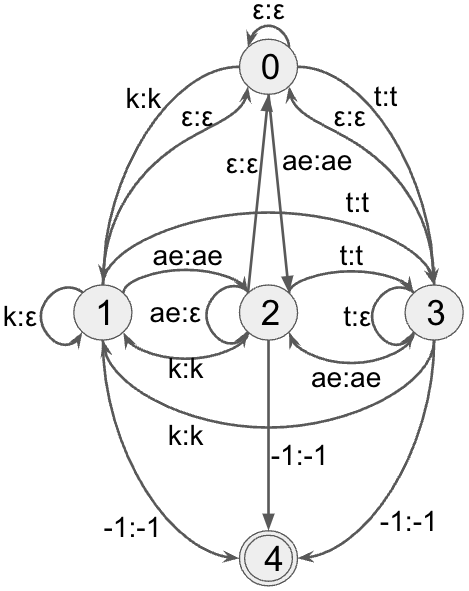}}
    \caption{$H$ topology of phone set \{k, ae, t\}. It merges duplicated adjacent phones in the input sequence, e.g., (k, k, ae, t, t) $\rightarrow$ (k, ae, t). The arc with $-1$ is a special arc defined in k2 pointing to the final state.}
    \label{fig:h}
\end{minipage}
\hfill
\begin{minipage}[b]{0.48\linewidth}
    \centering
    \centerline{\includegraphics[width=3.5cm]{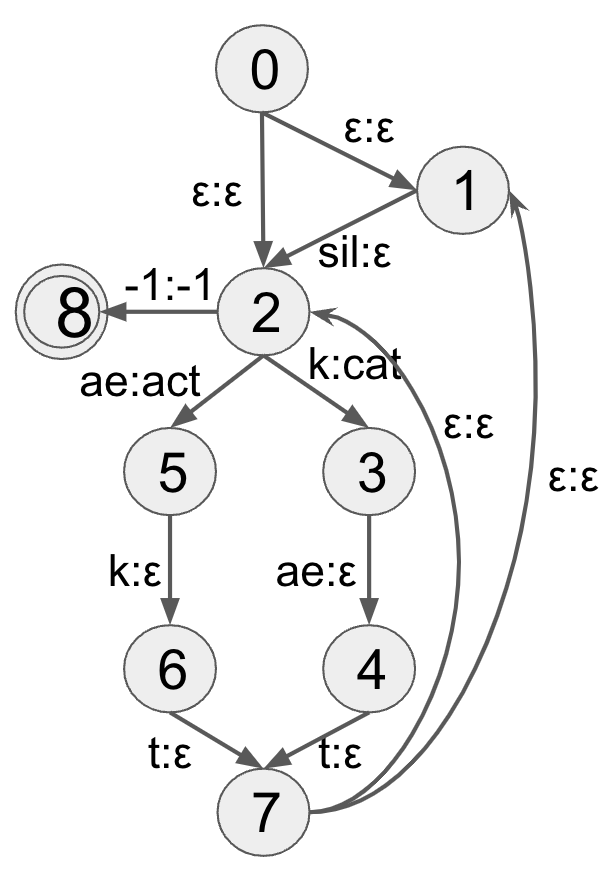}}
    %\vspace{-10pt}
    \caption{$L$ topology of lexicon \{cat: k, ae, t; act: ae, k, t\} in k2. It maps the phone sequence to the corresponding word allowing optional silence token (sil) between words. e.g., (sil, k, ae, t, sil) $\rightarrow$ cat.}
    \label{fig:lexicon}
\end{minipage}
\vspace{-15pt}
\end{figure}

As briefly introduced in Sec.~\ref{sec:intro}, the graph-based search employs WFST for decoding. Different from the prefix beam search method, the graph-based search can utilize word-level LMs in the search graph, and can also be extended to word recognition. The search graph $T$ is a composition of three functional graphs: an alignment graph $H$, a lexicon graph $L$, and a grammar graph $G$:

\begin{equation}
    T = H \circ L \circ G,
\end{equation}
where $\circ$ is WFST composition. Specifically, $H$ merges duplicated adjacent phones; $L$ maps sequences of phonemes to  sequences of responding words; $G$ is an n-gram word LM. Fig.~\ref{fig:h} and Fig.~\ref{fig:lexicon} show an example of $H$ and $L$ implemented in k2. The lattice is generated during decoding~\cite{lattcie}, which represents the set of most likely hypothesis transcripts structured in a directed graph and can be easily integrated with neural LMs by performing lattice scoring~\cite{rnnlm_rescoring, parallel_rescoring}. The best hypothesis is obtained by searching the best path in the lattice.

\subsection{Recipes for reproducible experiments}
\label{ssec: recipes}

\subsubsection{Directory structure}

EURO follows the unified directory organization of ESPnet as shown in Fig.~\ref{fig:tree}. Similar to other tasks (e.g., ASR, speech translation (ST)), the recipe \texttt{usar.sh} and its related bash scripts are stored under \texttt{egs2/TEMPLATE/uasr/}. 
The model \texttt{espnet\_model.py} and task \texttt{uasr.py} are stored at \texttt{espnet2/uasr/} and \\ \texttt{espnet2/tasks/}, respectively. Decoding scripts \\ \texttt{uasr\_inference.py} and 
\texttt{uasr\_inference\_k2.py} are placed under \texttt{espnet2/bin/}.

\subsubsection{Recipe flow}

The recipe in EURO follows the ESPnet2 style of task, which provides the template \texttt{uasr.sh}. The stages are defined as follows:

\noindent \textit{\textbf{Stage 1-5}: Data preparation.} The initial data format starts from the Kaldi style \cite{kaldi}, but the \texttt{text} for transcription is supposed to be unpaired with the speech data. Then, we offer two optional data preprocessing stages: speed perturbation and voice activity detection (VAD). The VAD results can be applied in silence removal as it is shown to be important for some speech corpus for wav2vec-U. After the preprocessing, all speech data is converted into a standard format, by resampling, segmentation, silence removal, and dumping from pipe-style formats. \\
\noindent \textit{\textbf{Stage 6-7}: Text tokenization and token list generation.} Texts are converted to phoneme tokens using the graphemes to phonemes toolkit g2p-en. Tokens are collected from the training text and formed into a corresponding token list for modeling. For UASR, the unpaired text~$\mathbf{Y}_u$ is fed into training in a random fashion. For efficiency purposes,  we initialize a randomized text loader with the tokenized text, which is especially useful for large text data. \\ 
\noindent \textit{\textbf{Stage 8-11}: LM preparation.} These stages train and evaluate LMs based on the unpaired text $\mathcal{D}_{\text{text}}$. The LMs include both neural-based LMs and N-gram LM. \\ 
\noindent \textit{\textbf{Stage 12}: WFST graph construction.} This stage creates the WFST decoding graph for the k2 decoder.\\
\noindent \textit{\textbf{Stage 13}: UASR statistics collection.} In this stage, EURO collects necessary input statistics for batching and mean-variance normalization (MVN). Optionally, the feature from frontends (i.e., S3PRL module in EURO) can be extracted at the stage to support efficient training in further stages. \\ 
\noindent \textit{\textbf{Stage 14}: UASR feature preprocessing.} This stage applies the preprocessing steps discussed in Sec.~\ref{ssec: models}, including adjacent clustering pooling, PCA, and mean pooling. The resulting segments are used for UASR training. \\ 
\noindent \textit{\textbf{Stage 15}: UASR training.} This stage conducts the adversarial training as discussed in Sec.~\ref{ssec: models}. \\ 
\noindent \textit{\textbf{Stage 16}: UASR decoding.} EURO has two decoding schemes as introduced in Sec.~\ref{ssec: decoding}. This stage supports both decoding methods. \\
\noindent \textit{\textbf{Stage 17}: UASR evaluation.} The evaluation in this stage utilizes the NIST \textit{sclite} toolkit to compute the phone error rate (PER) or the word error rate (if applicable). \\ 
\noindent \textit{\textbf{Stage 18-20}: Model packing and uploading.} This stage automatically packs the trained model checkpoint for easier sharing of the pre-trained model. EURO also supports uploading models to Huggingface for model sharing.

%\begin{table}[t]
%\centering
%\caption{Details of SSL models used for experiments.}
%\label{lab:ssl models}
%\vspace{2mm}
%\begin{tabular}{llll}
%\hline
%Model                  & \#Layers    & Pre-trained data                                                                                                   \\ \hline
%wav2vec2\_large\_ll60k & 24  & 60k hours of Libri-Light                                                                                           \\ \hline
%hubert\_large\_ll60k   & 24   & 60k hours of Libri-Light                                                                                           \\ \hline
%wavlm\_large           & 24    & \begin{tabular}[c]{@{}l@{}}60k hours of Libri-Light\\ 10k hours of GigaSpeech~\cite{gigaspeech}\\ 24khours of VoxPopuli~\cite{vox}\end{tabular} \\ \hline
%\end{tabular}
%\end{table}

\begin{table*}[t]
\centering
\caption{PER ($\%$) on TIMIT of different SSL models. SE-ODM~\cite{sedom} only reports results on the test set. The best result is highlighted in \textbf{bold}. Experimental details can be found in Sec.~\ref{ssec: exp details}.}
\vspace{1mm}
\label{tab: timit}
\begin{tabular}{@{\extracolsep{4pt}}llllllll@{}}
\hline
\multirow{2}{*}{Framework} & \multirow{2}{*}{SSL model} & \multirow{2}{*}{Layer} & \multirow{2}{*}{LM} & \multicolumn{2}{c}{TIMIT} \\ \cline{5-6} 
                           &                            &                        &                     & dev            & test                   \\ \hline 
 SE-ODM \cite{sedom}                       & -                          & -                      & 5-gram              &  -             & 36.5                     \\ \hline
wav2vec-U \cite{wav2vecu}                  & wav2vec 2.0                & 15                     & 4-gram              & 17.0           & 17.8                           \\ \hline
\multirow{3}{*}{EURO}      & wav2vec 2.0                & 15                     & 4-gram              & 18.5           & 19.8                            \\ 
                           & HuBERT                     & 15                     & 4-gram              & 14.9           & 16.4                        \\
                           & WavLM                      & 14                     & 4-gram              & \textbf{14.3}  & \textbf{14.6}                        \\ 
\hline
\end{tabular}
\end{table*}

\begin{table*}[t]
\centering
\caption{PER/WER ($\%$) on Librispeech of different SSL models. PER is from the prefix beam search decoder and WER is from the k2 WFST decoder. The best result is highlighted in \textbf{bold}. Experimental details can be found in Sec.~\ref{ssec: exp details}.}
\vspace{1mm}
\label{tab: librispeech}
\begin{tabular}{@{\extracolsep{4pt}}llllllllll@{}}

\hline
\multirow{2}{*}{Framework} & \multirow{2}{*}{SSL model} & \multirow{2}{*}{Layer} & \multirow{2}{*}{LM} & \multicolumn{4}{c}{Librispeech} \\ \cline{5-8}
                           &                            &                        &                     & dev clean  & dev other   & test clean &test other                   \\ \hline
wav2vec-U  & wav2vec 2.0            & 15                 & 4-gram              & 18.9/31.7       & 22.4/35.4         & 18.4/30.7    & 23.0/36.1             \\ \hline
\multirow{3}{*}{EURO}      & wav2vec 2.0            & 15                 & 4-gram              & 16.2/26.5       & \textbf{19.3}/29.8         & 15.7/25.6    & \textbf{19.8}/30.7       \\ 
                           & HuBERT                 & 15                 & 4-gram              & \textbf{15.2}/23.1       & 20.7/\textbf{29.3}   & \textbf{15.1}/\textbf{22.8}    & 21.1/\textbf{29.8}        \\
                           & WavLM                  & 15                 & 4-gram              & 18.0/\textbf{23.0}           & 21.2/31.0      & 16.6/22.9        & 21.4/31.0        \\ 
\hline
\end{tabular}

\end{table*}

\section{Experiments}
\label{sec:experiments}
\vspace{-3mm}
\subsection{Datasets}
\noindent \textbf{TIMIT}: The TIMIT dataset is a popular benchmark for the UASR task \cite{timit}. It contains 6300 sentences (5.4 hours) of reading speech. All sentences are manually transcribed to phonemes with time alignment. We use the standard split of the train (3696 sentences), dev (400 sentences), and test (192 sentences) sets for our experiments.\\

\noindent \textbf{Librispeech}: The LibriSpeech dataset is a common benchmark for the ASR task~\cite{librispeech}. It contains 960 hours of reading speech automatically derived from the audiobooks of LibriVox. This corpus is split into 3 training sets (100 and 360 hours of clean speech, and 500 hours of other speech), 2 dev sets (each has 5 hours), and test sets (each has 5 hours).

%\subsubsection{Totonac}
%TODO(Jiatong)
\vspace{-2mm}
\subsection{SSL models}
\vspace{-2mm}
We explore three SSL models for UASR in EURO, wav2vec 2.0 (wav2vec2-large-ll60k), HuBERT (hubert-large-ll60k), and WavLM (wavlm-large).\footnote{Corresponding models can be found in \url{https://s3prl.github.io/s3prl/tutorial/upstream_collection.html}} The three models have the same architecture that consists of 24 layers of Transformer encoders~\cite{transformer} with a similar number of parameters. Among them, wav2vec 2.0 and HuBERT are pre-trained on 60,000 hours of Libri-Light. In addition to Libri-Light~\cite{kahn2020libri}, Wavlm uses 10,000 hours of Gigaspeech~\cite{gigaspeech} and 24,000 hours of VoxPopuli~\cite{vox} for pre-training.

\vspace{-2mm}
\subsection{Comparison of SSL models}

\noindent \textbf{TIMIT}: We test EURO on TIMIT to confirm that the toolkit works properly. We use the text from the same training set for unsupervised training. To make it comparable with wav2vec-U, we adopt the same setup for EURO wav2vec 2.0. More specifically, we use the model wav2vec2-large-ll60k and the features are extracted from the \nth{15} layer of the model. For Hubert and WavLM, we explore the performance of features from different layers and report the best PER results. 

Table~\ref{tab: timit} shows the results on TIMIT. Wav2vec-U serves as the baseline and achieves 17.0\% and 17.8\% PER on dev and test sets, respectively. 
EURO with the same setup gets 18.5\% and 19.8\% PER on the dev and test set which is slightly worse than wav2vec-U. While our model is not heavily tuned under this setup, the results are comparable with wav2vec-U.  
For Hubert (hubert-large-ll60k), EURO performs best on features extracted from the \nth{15} layer. The PERs on the dev set and test set are 14.9\% and 16.4\%. Compared with the wav2vec-U baseline and EURO wav2vec 2.0, it provides a relative improvement of 10\% and 20\%, respectively. 
WavLM (wavlm-large) provides further improvement. The model trained using features extracted from \nth{14} layer of WavLM achieves 17\% relative improvement compared with baseline and 25\% relative improvement compared with EURO wav2vec 2.0 in terms of PER. It reduces the PER to 14.3\% on the dev set and 14.6\% on the test set.\\
\\
\noindent \textbf{Librispeech}: For Librispeech, because of the limitation of computing resources, we use 100 hours of clean speech for UASR training. Unlike TIMIT, we use the text from the whole training set (960 hours) excluding the overlap part with the training speech. The text is phonemicized using G2P phonemizer~\cite{g2p}. In total, around 25m sentences are used for UASR training. 

We measure both the PER and WER of UASR systems using different SSL models for this dataset. Wav2vec-U uses a sophisticated decoder to convert phoneme sequences to word sequences and it may not be easily applied to other datasets. To make a fair comparison, we train the wav2vec-U model using the same data and load the model into EURO. We use the same prefix beam search decoder for PER and k2 WFST decoder for WER.
Table~\ref{tab: librispeech} show the results of LibriSpeech's standard dev and test sets. All EURO models outperform the baseline wav2vec-U. For phone recognition, the HuBERT model performs best on clean sets. It achieves PER 15.2 on the dev clean set and PER 15.1 on the test clean set. Wav2vec 2.0 performs best on more difficult sets. It achieves PER 19.3 and 19.8 on dev other and test other sets, respectively. For word recognition, Hubert gets the best WER of 29.3, 22.8, and 29.8 on dev other, test clean and test other sets. WavLM performs best on dev clean and achieves 23.0 WER which is slightly better than HuBERT.
%Overall, HuBERT model performs best on this dataset. It achieves best PER results on dev clean (15.2) and test clean sets (15.1), and best WER resuls on dev other (29.3) and all test sets (22.8 and 29.8, respectively). Wav2vec 2.0 performs better on the noisy dev other (19.3) and test other () sets for phone recognition. WavLM 

%\subsubsection{Results on Low-resource languages}
%TODO(Jiatong)
%% Please add the following required packages to your document preamble:
%% \usepackage{multirow}
%\begin{table}[]
%\begin{center}
%\caption{PER (\%) on Totonac of different SSL models. }
%\begin{tabular}{llll}
%\hline
%\multirow{2}{*}{}     & \multirow{2}{*}{SSL model} & \multicolumn{2}{c}{Totonac} \\ \cline{3-4} 
%                      &                            & dev          & test         \\ \hline
%\multirow{3}{*}{EURO} & wav2vec 2.0                &              &              \\
%                      & HuBERT                     &              &              \\
%                      & WavLM                      &              &              \\ \hline 
%\end{tabular}
%\end{center}
%\end{table}

%\subsection{Stabilities}
%The stability of all three SSL models are consistent in EURO.

\subsection{Experimental details}
\label{ssec: exp details}
For training, we set $\lambda=1.5$, $\gamma=0.5$, and $\eta=2.0$ for TIMIT and $\lambda=2.0, \gamma=1.0, \eta=4.0$ for LibriSpeech datasets.
For (phone-level) prefix beam search decoding, we set $\text{beam size}=2$ and tunes the weight n-gram language model in $\left[ 0,0.9 \right]$. For (word-level) graph-based WFST decoding, we set $\text{search beam size}=30$, $\text{output beam size}=15$, $\text{min active states}=14,000$, and $\text{max active stats}=56,000$. More details can be found in the configuration file under \texttt{egs2/TEMPLATE/uasr/conf/}.

%\begin{table}[h]
%\label{tab:exp details}
%\centering
%\caption{Hyperparameters: weights of gradient penalty loss, smoothness loss, and phone diversity loss}
%\begin{tabular}{l|l|l|l}
%\hline
%            & gp ($\lambda$) & sp ($\gamma$) & pd ($\eta$) \\ \hline
%TIMIT       & 1.5              & 0.5                & 2.0                  \\ \hline
%LibriSpeech & 2.0              & 1.0                & 4.0                  \\ \hline
%\end{tabular}
%\end{table}

%The paper title (on the first page) should begin 1.38 inches (35 mm) from the
%top edge of the page, centered, completely capitalized, and in Times 14-point,
%boldface type.  The authors' name(s) and affiliation(s) appear below the title
%in capital and lower case letters.  Papers with multiple authors and
%affiliations may require two or more lines for this information. Please note
%that papers should not be submitted blind; include the authors' names on the
%PDF.

\section{Conclusions}
\label{sec:conclusions}
This work introduces a new toolkit for unsupervised ASR, namely EURO. The toolkit is developed as an open platform for the research field of unsupervised ASR. The current architecture of EURO is based on the Wav2vec-U framework but greatly improves the reproducibility with flexible frontends of almost 30 SSL models and a faster preparation/inference compared to its original implementation in FAIRSEQ. By integrating with k2, EURO provides a WFST decoder for word recognition. Our experiments on TIMIT and LibriSpeech show that we could get comparable performances with wav2vec-U in FAIRSEQ but even better results with Hubert and WavLM as new frontends.

%To achieve the best rendering both in printed proceedings and electronic proceedings, we
%strongly encourage you to use Times-Roman font.  In addition, this will give
%the proceedings a more uniform look.  Use a font that is no smaller than nine
%point type throughout the paper, including figure captions.
%
%In nine point type font, capital letters are 2 mm high.  {\bf If you use the
%smallest point size, there should be no more than 3.2 lines/cm (8 lines/inch)
%vertically.}  This is a minimum spacing; 2.75 lines/cm (7 lines/inch) will make
%the paper much more readable.  Larger type sizes require correspondingly larger
%vertical spacing.  Please do not double-space your paper.  TrueType or
%Postscript Type 1 fonts are preferred.
%
%The first paragraph in each section should not be indented, but all the
%following paragraphs within the section should be indented as these paragraphs
%demonstrate.

\section{Acknowledgement}
\label{sec:acknowledgement}
Part of the work presented here was carried out during the 2022 Jelinek Memorial Summer Workshop on Speech and Language Technologies at Johns Hopkins University, which was supported with unrestricted gifts from Amazon, Microsoft, and Google. This work used the Bridges system ~\cite{nystrom2015bridges}, which is supported by NSF award number ACI-1445606, at the Pittsburgh Supercomputing Center (PSC).
%Major headings, for example, "1. Introduction", should appear in all capital
%letters, bold face if possible, centered in the column, with one blank line
%before, and one blank line after. Use a period (".") after the heading number,
%not a colon.

%Subheadings should appear in lower case (initial word capitalized) in
%boldface.  They should start at the left margin on a separate line.

%Sub-subheadings, as in this paragraph, are discouraged. However, if you
%must use them, they should appear in lower case (initial word
%capitalized) and start at the left margin on a separate line, with paragraph
%text beginning on the following line.  They should be in italics.

\clearpage

\bibliographystyle{IEEEbib}
\bibliography{strings,refs}

\end{document}